\documentclass[runningheads]{llncs}

\usepackage{graphicx}
\usepackage{amsmath}
\usepackage{amssymb}
\usepackage{booktabs}
\usepackage{multirow}
\usepackage{algorithm}
\usepackage{algorithmic}
\usepackage{cite}

\begin{document}

\title{NeoNet: An End-to-End 3D MRI-Based Deep Learning Framework for Non-Invasive Prediction of Perineural Invasion via Generation-Driven Classification}

\titlerunning{NeoNet: 3D MRI-Based Prediction of PNI}

\author{
Youngung Han\inst{1,2}
\and Minkyung Cha\inst{1}
\and Kyeonghun Kim\inst{2}
\and Induk Um\inst{3}
\and Myeongbin Sho\inst{4}
\and Joo Young Bae\inst{1}
\and Jaewon Jung\inst{1}
\and Jung Hyeok Park\inst{1}
\and Seojun Lee\inst{1}
\and Nam-Joon Kim\inst{1,}\thanks{Corresponding Author}
\and Woo Kyoung Jeong\inst{5}
\and Won Jae Lee\inst{6}
\and Pa Hong\inst{7}
\and Ken Ying-Kai Liao\inst{8}
\and Hyuk-Jae Lee\inst{1}
}

\authorrunning{Han et al.}

\institute{
Seoul National University, South Korea\\
\email{yuhan@snu.ac.kr, \textsuperscript{*}knj01@snu.ac.kr}
\and OUTTA, South Korea\\
\and Chung-Ang University, South Korea\\
\and Sookmyung Women's University, South Korea\\
\and Samsung Medical Center, Sungkyunkwan University School of Medicine, Seoul, Republic of Korea\\
\and Samsung Changwon Hospital, Sungkyunkwan University School of Medicine, Changwon, Republic of Korea\\
\and Samsung Changwon Hospital, Sungkyunkwan University School of Medicine\\
\and NVIDIA AI Technology Center, Taipei, Taiwan
}

\maketitle

\begin{abstract}
Minimizing invasive diagnostic procedures to reduce the risk of patient injury and infection is a central goal in medical imaging. And yet, noninvasive diagnosis of perineural invasion (PNI), a critical prognostic factor involving infiltration of tumor cells along the surrounding nerve, still remains challenging, due to the lack of clear and consistent imaging criteria criteria for identifying PNI. To address this challenge, we present NeoNet, an integrated end-to-end 3D deep learning framework for PNI prediction in cholangiocarcinoma that does not rely on predefined image features. NeoNet integrates three modules: (1) NeoSeg, utilizing a Tumor-Localized ROI Crop (TLCR) algorithm; (2) NeoGen, a 3D Latent Diffusion Model (LDM) with ControlNet, conditioned on anatomical masks to generate synthetic image patches, specifically balancing the dataset to a 1:1 ratio; and (3) NeoCls, the final prediction module. For NeoCls, we developed the PNI-Attention Network (PattenNet), which uses the frozen LDM encoder and specialized 3D Dual Attention Blocks (DAB) designed to detect subtle intensity variations and spatial patterns indicative of PNI. In 5-fold cross-validation, NeoNet outperformed baseline 3D models and achieved the highest performance with a maximum AUC of 0.7903.
\end{abstract}

\section{Introduction}
Nerves throughout the human body are organized into bundles of fibers surrounded by the perineurium, a protective connective tissue sheath. When cancer cells infiltrate this structure and spread in or along the nerves, the process is referred to as perineural invasion (PNI)~\cite{Liebig2009}.

PNI is recognized as an active process facilitated by biochemical crosstalk between tumor cells and the nerve~\cite{Amit2016}, serving as an independent route of metastasis. It is a critical prognostic factor, particularly in cholangiocarcinoma, where patients with PNI show significantly higher recurrence rates after surgery~\cite{Hruban2022, Qian2024}. Accurate preoperative identification of PNI is therefore of significant clinical value, as it enables better risk stratification, informs surgical planning, and could act as a guide for clinical decision-making for adjuvant therapy.

\begin{figure}[t]
\centering
\includegraphics[width=0.85\columnwidth]{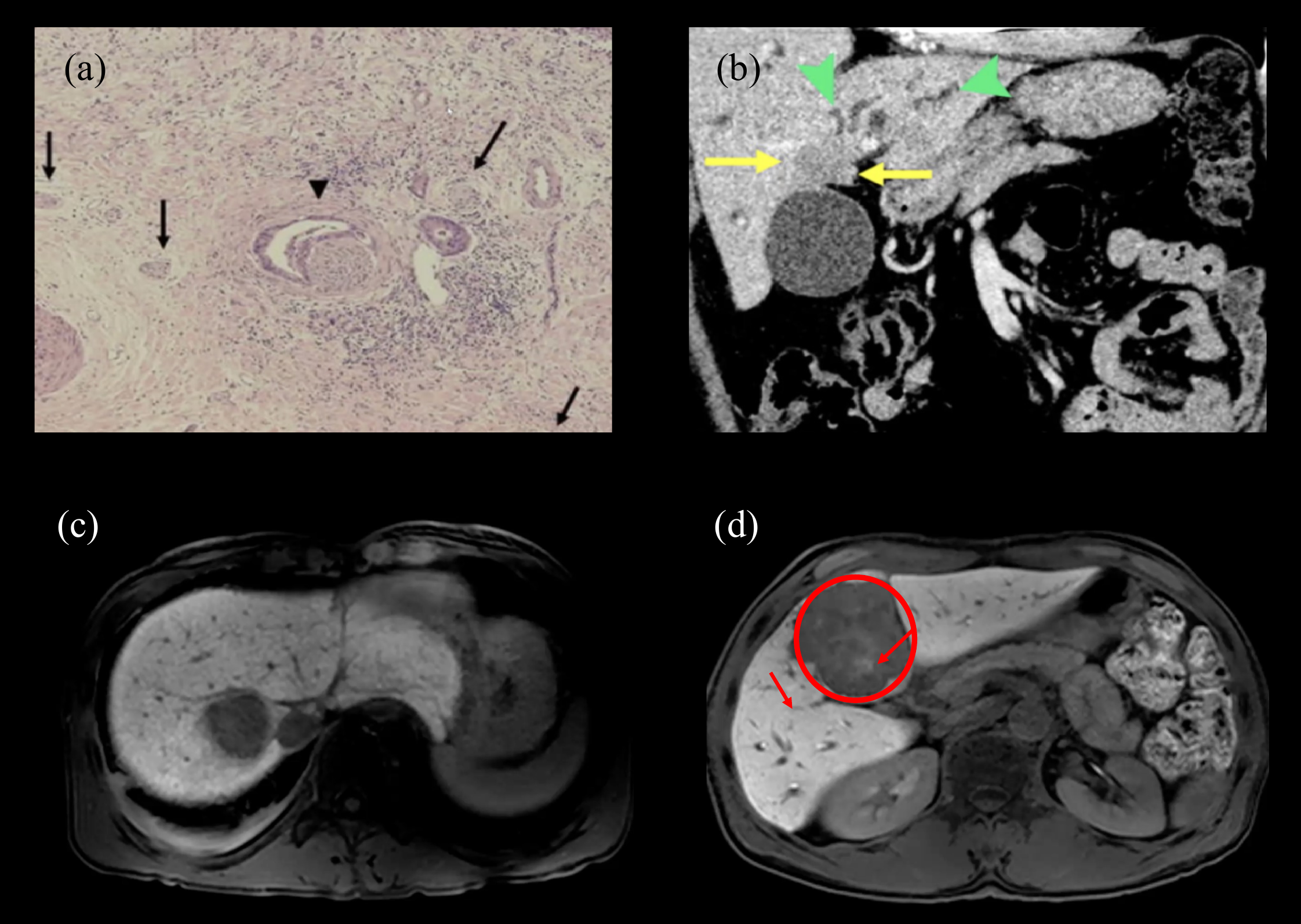}
\caption{Morphological features of perineural invasion (PNI) demonstrated by histopathology (a), CT (b), and MRI (c-PNI negative, d-PNI positive). Arrows indicate the locations of characteristic PNI features in each modality.}
\label{fig1}
\end{figure}

Despite the need, efforts to predict PNI non-invasively have remained challenging. A key reason lies in the intrinsic difficulty of accurately identifying PNI through conventional imaging. While features of PNI are relatively clear in post-surgical histopathology images, corresponding signs on CT and MRI are often subtle and less definitive (Figure~\ref{fig1}), appearing as minor nerve thickening or heterogeneous enhancement~\cite{Purohit2019}.

Compounding this challenge is the profound scarcity of well-labeled PNI datasets. At present, Definitive PNI identification requires postoperative histopathological confirmation, meaning that only MRI scans from surgically resected cases can be reliably labeled, thereby constraining data availability. The scarcity of data is a field-wide issue; some related studies rely on small cohorts typically fewer than 200 cases \cite{Huang2021, Qi2025}.

Methodologically, current approaches predominantly rely on radiomics~\cite{Gillies2016} or handcrafted features and manual segmentation \cite{Huang2021}. Even recent advancements incorporating deep learning often utilize it as part of a hybrid approach rather than a fully end-to-end framework \cite{Qi2025}. 

Therefore, we propose NeoNet, an integrated 3D deep learning framework designed to overcome prior limitations. Our framework is the first end-to-end 3D deep learning approach for PNI prediction that integrates automated localization, generative data balancing, and specialized classification. By moving beyond traditional radiomics and addressing data scarcity head-on, NeoNet aims to provide a non-invasive clinical tool for preoperative identification of PNI.

Our main contributions are summarized as follows:
\begin{itemize}
    \item We propose NeoNet, the first integrated end-to-end 3D deep learning framework specifically designed for PNI prediction from MRI, overcoming the limitations of radiomics-based approaches by learning features directly from the imaging data.
    \item We utilize NeoGen, a 3D LDM with ControlNet, to address the critical challenges of data scarcity and severe class imbalance inherent in PNI datasets by generating anatomically constrained synthetic data and balancing the training cohort to a 1:1 ratio.
    \item We design PattenNet, a lightweight 3D classifier incorporating a frozen LDM encoder and Dual Attention Blocks (DAB) specifically engineered to detect the subtle, localized features of PNI by focusing on critical intensity variations and spatial patterns at the tumor interface.
\end{itemize}

\section{Related Works}

\subsubsection{PNI Prediction in Medical Imaging}
The non-invasive prediction of PNI is challenging due to the subtlety of its features on conventional imaging. Traditional assessment relies on subjective radiological interpretation of MRI features~\cite{Purohit2019}.

To achieve quantitative assessment, radiomics~\cite{Gillies2016, Lambin2017} has emerged as the predominant approach. Methodologically, however, radiomics relies on handcrafted features extracted from predefined regions of interest, often requiring labor-intensive manual segmentation. 

Data scarcity is an inherent obstacle, as definitive PNI diagnosis requires post-surgical pathology. Consequently, publicly available datasets are nonexistent, and studies rely on small, private cohorts. For instance, \cite{Huang2021} utilized a cohort of 101 patients to predict PNI in extrahepatic cholangiocarcinoma using radiomics features and machine learning classifiers. These small datasets invariably suffer from significant class imbalance, typically addressed using statistical oversampling (e.g., SMOTE), which may not be optimal for high-dimensional imaging data.

The adoption of end-to-end deep learning has been severely limited by these constraints. Small datasets hinder the ability of deep networks to learn generalizable features. Recent work by \cite{Qi2025} utilized a larger cohort of 192 cases but employed a hybrid fusion model combining ResNet101~\cite{He2016} features, radiomics, and clinical data, rather than a pure end-to-end deep learning approach for feature learning.


\subsubsection{Localization in Medical Image Classification}
For localized pathologies like PNI, which manifest at the tumor-nerve interface, focusing the model's attention on the relevant anatomical region is crucial. While end-to-end frameworks combining segmentation and classification are common in medical imaging, the precise localization strategy is vital for performance. 

\subsubsection{Generative Models for Medical Data Augmentation}
Data scarcity and class imbalance are major challenges in medical deep learning. Generative models offer a powerful tool for data augmentation and balancing~\cite{Yi2019, Kazeminia2020}. Variational AutoEncoders (VAEs)~\cite{Kingma2013} and Generative Adversarial Networks (GANs)~\cite{Goodfellow2014} have been widely used. More recently, Diffusion models~\cite{Sohl-Dickstein2015} have shown remarkable success in high-fidelity image generation, including DDPM~\cite{Ho2020} and Latent Diffusion Models (LDM)~\cite{Rombach2022}. LDM operates in a compressed latent space, making it efficient for high-resolution 3D data, and has shown promise in medical domains such as brain imaging~\cite{Pinaya2022}.

To control the generation process, conditional mechanisms are crucial. ControlNet~\cite{Zhang2023} allows for integrating external control signals, such as anatomical masks. In this work, we leverage a 3D LDM (NeoGen) with ControlNet to generate realistic synthetic MRI patches conditioned on segmentation masks, which are then used to balance the training dataset.

\begin{figure*}[t]
\centering
\includegraphics[width=1.0\textwidth]{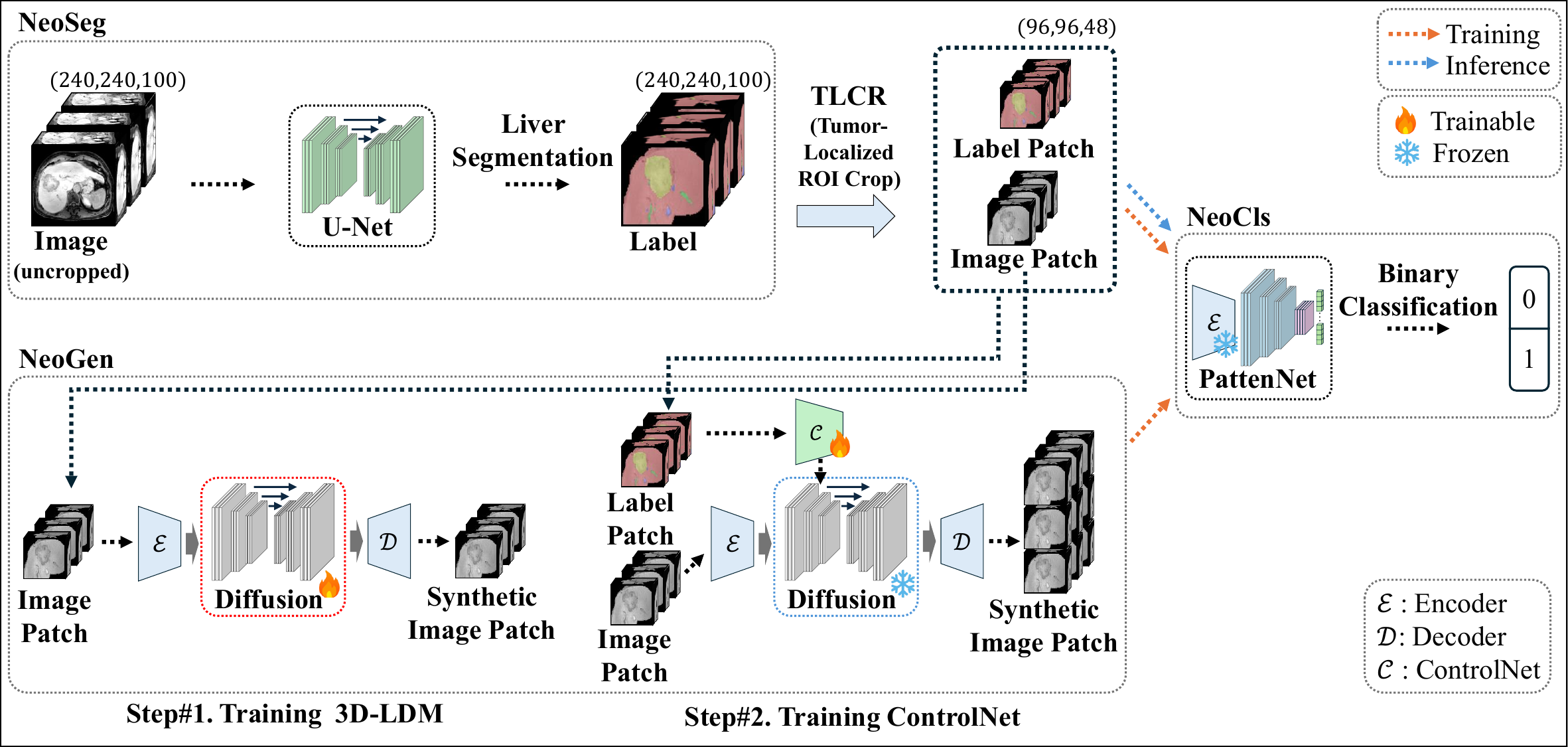}
\caption{Overview of the NeoNet framework. (1) NeoSeg performs automated liver and tumor segmentation on inputs (e.g., $240\times 240\times 100$), producing segmentation labels (e.g., $240\times 240\times 100$). The TLCR algorithm utilizes these labels and the original image to extract localized image and label patches ($96\times 96\times 48$). (2) NeoGen utilizes a 3D-LDM (Step \#1) and trains ControlNet (Step \#2) conditioned on label patches to generate synthetic image patches, balancing the dataset. (3) NeoCls utilizes PattenNet, initialized with the frozen LDM encoder ($\mathcal{E}$), to predict PNI status from real and synthetic patches.}
\label{fig3}
\end{figure*}

\section{Methodology}

NeoNet (Figure~\ref{fig3}) is an integrated framework composed of three main modules: NeoSeg for segmentation and localization, NeoGen for synthetic data generation, and NeoCls for PNI prediction.

\subsection{Datasets and Preprocessing}
We utilized T1-weighted hepatobiliary phase MRI scans collected over a decade at Samsung Medical Center (SMC). An initial pool of 306 patients was manually inspected. 178 patients were excluded due to poor image quality, inaccurate annotations, incomplete sequences, or tumors occupying more than 70\% of the liver volume.

The final cohort included a total of 128 MRI scans, 44 PNI(+) and 84 PNI(-).  Images were provided in NIfTI format, along with ground truth annotations. For NeoSeg training, Intensity was normalized between 0~1 and labels were mapped to: background (0), liver (1), and tumor (2). 

\subsection{Experimental Protocol}
We employed a stratified 5-fold cross-validation strategy consistently across the entire framework. In each fold, the data was split into 80\% training and 20\% validation sets at the patient level. All components, including the LDM pretraining (VAE and U-Net) and ControlNet training, were strictly performed using only the training data of the respective fold.


\subsection{NeoSeg and Tumor Localization}
NeoSeg performs automated segmentation of the liver and tumor. We evaluated four architectures: U-Net~\cite{Ronneberger2015}, SegResNet~\cite{Myronenko2018}, DynUNet~\cite{Isensee2021}, and SwinUNETR~\cite{Tang2022}. SwinUNETR, leveraging advancements in Vision Transformers~\cite{Dosovitskiy2020, Liu2021}, achieved the highest performance and was selected as the backbone.

\begin{algorithm}[H]
\caption{Tumor-Localized ROI Crop Algorithm}
\label{alg:algorithm}
\textbf{Input}:\\3D medical image $\mathbf{I}$, crop size $\mathbf{C} = (c_x, c_y, c_z)$,\\
label map $\mathbf{L} =\{0:background, \ 1:liver, \  2:tumor\}$\\
\textbf{Output}: \\
Cropped vectors\\
$\mathbf{T_1} \in \mathbb{R}^{c_x \times c_y \times c_z}$ (peritumoral mask)\\
$\mathbf{T_2} \in \mathbb{R}^{c_x \times c_y \times c_z}$ (tumor mask)
\begin{algorithmic}[1] 
\STATE Find tumor voxel coordinates: \\
$\mathcal{S} \gets \{ (x, y, z)\ |\ \mathbf{L}[x, y, z] = 2 \}$
\IF{$\mathcal{S}$ is empty}
\STATE $\mathbf{T_1, T_2} \gets \mathbf{0}^{c_x \times c_y \times c_z}$
\STATE \textbf{return} $\mathbf{T_1, T_2}$
\ENDIF
\STATE $[x_{\min}, y_{\min}, z_{\min}] \gets \min(\mathcal{S})$\\
$[x_{\max}, y_{\max}, z_{\max}] \gets \max(\mathcal{S}) + 1$
\STATE $\mathbf{center} \gets \left( \left\lfloor \frac{x_{\min} + x_{\max}}{2} \right\rfloor, \left\lfloor \frac{y_{\min} + y_{\max}}{2} \right\rfloor, \left\lfloor \frac{z_{\min} + z_{\max}}{2} \right\rfloor \right)$
\STATE $\mathbf{start} \gets \max\left(\mathbf{center} - \frac{\mathbf{C}}{2}, 0\right)$\\ $\mathbf{end} \gets \mathbf{start} + \mathbf{C}$

\STATE Extract crop:\\ $\mathbf{I}_{crop} \gets \mathbf{I}[\mathbf{start} : \mathbf{start} + \mathbf{C}]$\\
$\mathbf{L}_{crop} \gets \mathbf{L}[\mathbf{start} : \mathbf{start} + \mathbf{C}]$
\STATE $\mathbf{T_1} \gets \mathbf{I}_{crop}[ \mathbf{L}_{crop} \in \{1, 2\}]$ \\
$\mathbf{T_2} \gets \mathbf{I}_{crop}[\mathbf{L}_{crop} \in \{2\}]$
\STATE \textbf{return} $\mathbf{T_1,T_2}$
\end{algorithmic}
\end{algorithm}

\subsubsection{Tumor-Localized ROI Crop (TLCR)}
Based on the segmentation results, we implement the TLCR algorithm (Algorithm~\ref{alg:algorithm}) to extract a standardized input volume focused on the tumor region. This approach is based on the premise that valid features for PNI prediction are localized in and around the tumor boundary.

TLCR calculates the geometric center of the tumor mask and extracts a fixed-size 3D crop of ($96\times96\times48$) voxels. The algorithm includes boundary clamping.

The output is a dual-channel volume: Channel 1 contains the peritumoral region (liver tissue and tumor), and Channel 2 contains only the tumor core. Voxels outside these masks are zeroed out. This dual-channel representation guides the classifier to focus on both the tumor characteristics and the tumor-liver interface.


\subsection{NeoGen: Conditioned Synthetic Generation}
To address data scarcity, NeoGen generates synthetic 3D MRI patches using a 3D Latent Diffusion Model (3D-LDM)~\cite{Rombach2022} enhanced with ControlNet~\cite{Zhang2023}.

\subsubsection{Addressing Data Scarcity and Imbalance}
The limited size of the PNI cohort of 128 cases and the inherent class imbalance pose significant challenges, especially when the absolute number of dataset is so small. NeoGen is utilized to mitigate these issues. By generating high-fidelity synthetic patches conditioned on anatomical masks, we augment the training dataset within each cross-validation fold, and generate sufficient synthetic samples to achieve a balanced 1:1 ratio between PNI-positive and PNI-negative samples in the final augmented training set used for NeoCls.

\begin{figure*}[t]
\centering
\includegraphics[width=1.0\textwidth]{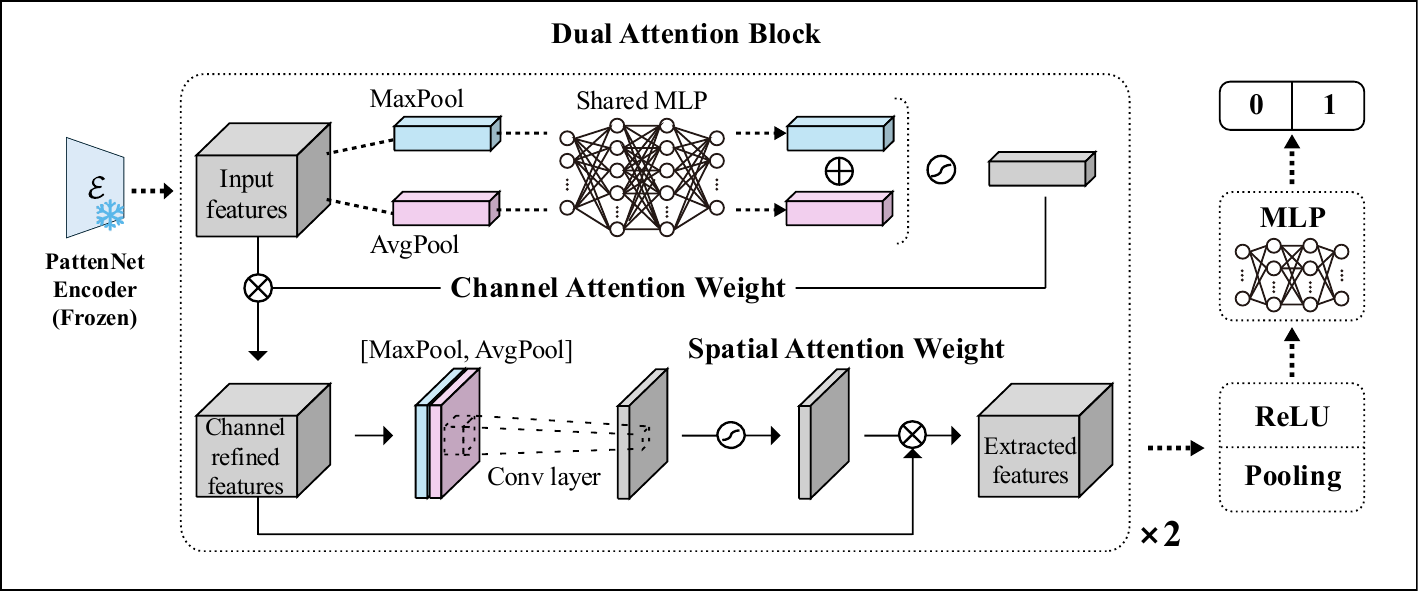}
\caption{Architecture of PattenNet for PNI prediction. The model utilizes a frozen encoder from the 3D-LDM followed by two Dual Attention Blocks (DAB). Each DAB integrates 3D channel and spatial attention. Spatial attention utilizes a 3D convolution ($7\times7\times7$) to effectively model 3D spatial relationships. The attended features are processed through global pooling and an MLP for binary classification.}
\label{fig4}
\end{figure*}

\subsubsection{3D Latent Diffusion Model}
The 3D-LDM operates in a compressed latent space learned by an autoencoder (VAE). The encoder $\mathcal{E}$ transforms a medical volume $x$ into a latent representation $z = \mathcal{E}(x)$. A diffusion model (3D U-Net) $\epsilon_{\theta}$ is trained to denoise the latent features $z_t$ at timestep $t$. The loss function is:
\begin{equation}
\mathcal{L}_{LDM} = \mathbb{E}_{\mathcal{E}(x), \epsilon \sim \mathcal{N}(0,1), t}[||\epsilon-\epsilon_{\theta}(z_{t},t)||_{2}^{2}]
\end{equation}

\subsubsection{ControlNet Conditioning}
ControlNet introduces trainable copies of the LDM blocks to integrate external control signals. Given a network block $F(\mathbf{x}; \theta)$, the controlled output $\mathbf{y}_c$ is:
\begin{equation}
\mathbf{y}_c = F(\mathbf{x}; \theta) + Z_2 \left( F\big( \mathbf{x} + Z_1(\mathbf{c}); \theta' \big) \right)
\end{equation}
where $\theta'$ are trainable parameters, and $Z_1, Z_2$ are zero-initialized $1\times1\times1$ convolution layers.

We employ the anatomical mask as the conditioning input ($\mathbf{c}$). The dual-channel label patch from TLCR guides the spatial structure of the generated image, ensuring anatomical plausibility.

The overall learning objective incorporates this condition:
\begin{equation}
\mathcal{L} = \mathbb{E}_{\mathbf{z}_0, \mathbf{c}, \epsilon, t} \left[ \left\| \epsilon - \epsilon_\theta(\mathbf{z}_t, t, \mathbf{c}) \right\|_2^2 \right]
\end{equation}

\subsection{NeoCls: PNI-Attention Network (PattenNet)}
NeoCls utilizes PattenNet (Figure~\ref{fig4}), a lightweight 3D classifier optimized for detecting subtle PNI features. PattenNet leverages the encoder from the trained 3D-LDM as a frozen feature extractor, followed by a stack of two Dual Attention Blocks (DABs).

\subsubsection{Dual Attention Block (DAB)}
To effectively capture the subtle intensity variations and complex spatial patterns characteristic of PNI, we designed the Dual Attention Block (DAB). DAB sequentially applies 3D channel and spatial attention mechanisms, adapted from concepts in CBAM~\cite{Woo2018} and general attention principles~\cite{Vaswani2017}, to refine the feature maps extracted by the frozen LDM encoder.

\textbf{Channel Attention} identifies what features are relevant to PNI by adaptively recalibrating channel-wise feature responses. For an input feature map $\mathbf{F} \in \mathbb{R}^{D\times H\times W\times C}$, 3D global max and average pooling are applied, followed by a shared MLP:
\begin{equation}
\mathbf{M}_{c}(\mathbf{F}) = \sigma(\text{MLP}(\text{AvgPool}(\mathbf{F})) +\text{MLP}(\text{MaxPool}(\mathbf{F})))
\end{equation}

\textbf{3D Spatial Attention} identifies where the PNI features are located, focusing attention on critical regions such as the tumor-liver interface. Pooling is applied across the channel dimension, and the resulting map is processed by a 3D convolution layer:
\begin{equation}
\mathbf{M}_{s}(\mathbf{F}) = \sigma(f^{7\times 7\times 7}([\text{AvgPool}_c(\mathbf{F});\text{MaxPool}_c(\mathbf{F})]))
\end{equation}
where $f^{7\times 7\times 7}$ denotes a 3D convolution with a $7\times 7\times 7$ kernel to capture sufficient 3D spatial context.

The refined feature map is obtained by sequentially applying the attention maps. After the DABs, the features are passed through global pooling and a fully connected layer for binary classification.

\subsection{Training Protocol and Implementation Details}
All modules were trained strictly following the 5-fold CV protocol defined previously. All training was performed on NVIDIA A100 GPUs, utilizing the AdamW optimizer~\cite{Loshchilov2017} consistently across all stages.

\subsubsection{NeoSeg}
NeoSeg (SwinUNETR) training utilized a combined Dice and Cross-Entropy loss, with a learning rate (LR) of $10^{-4}$ for 400 epochs with a batch size of 2.

\subsubsection{NeoGen}
The LDM training involved two stages within each fold.
\textbf{Stage 1 (LDM Training):} The VAE and 3D U-Net were trained from scratch using the fold's training split. The VAE was trained for 6,000 epochs with L1 loss, KL regularization of $10^{-7}$, and a learning rate of $10^{-6}$. The 3D U-Net was trained for 6,000 epochs with L2 loss and a learning rate of $10^{-5}$. \textbf{Stage 2 (ControlNet Training):} ControlNet was trained for 3,000 epochs, keeping the LDM weights frozen. It was optimized using MSE loss with a learning rate of $10^{-5}$ and batch size of 8, conditioned on the anatomical masks. Synthetic image patches were generated post-training to balance the dataset.

\subsubsection{NeoCls}
PattenNet and baseline 3D models, which include ResNet~\cite{He2016}, DenseNet~\cite{Huang2017}, EfficientNet~\cite{Tan2019}, and SwinTransformer~\cite{Liu2021}, were trained using the dual-channel TLCR outputs. We evaluated models trained on the original imbalanced data (R) and models trained on the augmented dataset (R+S). For R+S training, NeoGen was utilized to generate synthetic data until a balanced 1:1 ratio was achieved.

Training utilized a learning rate of $10^{-4}$ with batch size 4 for up to 300 epochs, with early stopping that had patience of 20 epochs based on validation AUC. The LDM encoder weights in PattenNet, which were trained in Stage 1 of the respective fold, were kept frozen.

\subsection{Evaluation Metrics}
Segmentation performance was evaluated using the Dice Similarity Coefficient (Dice). Generation quality was evaluated using Fréchet Inception Distance (FID)~\cite{Heusel2017}, PSNR, SSIM, and LPIPS~\cite{Zhang2018}. Classification performance was evaluated using the Area Under the Receiver Operating Characteristic Curve (AUC), reporting the mean AUC across the 5-fold cross-validation.

\section{Experiments}

\subsection{Liver and Tumor Segmentation Performance}
\begin{figure}[ht]
\centering
\includegraphics[width=0.85\columnwidth]{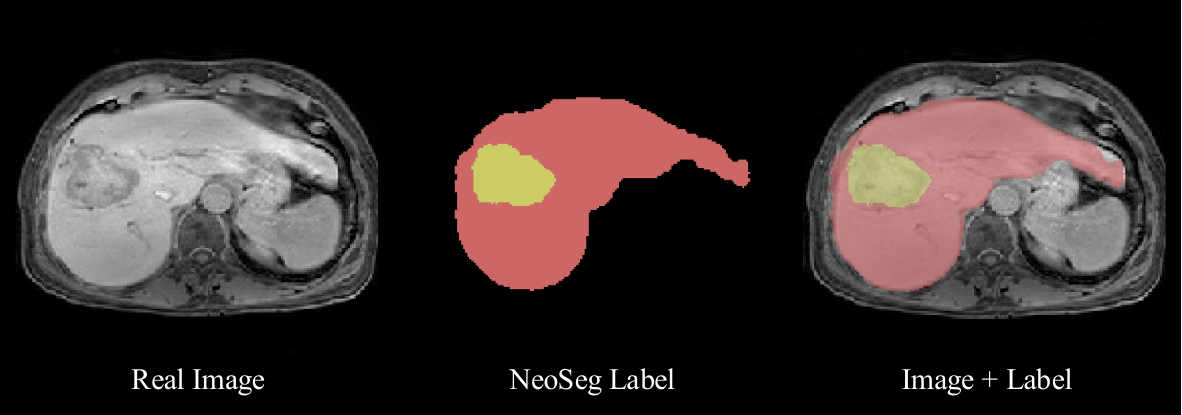}
\caption{Labels generated by NeoSeg. Real Image (left), Segmentation masks (middle: peritumoral mask in red, tumor in yellow), and Overlay (right).}
\label{fig5}
\end{figure}

Table~\ref{tab:seg_performance} presents the segmentation performance. SwinUNETR achieved the highest mean Dice score (0.9516) and was selected as the NeoSeg backbone. Figure~\ref{fig5} shows representative segmentation results, illustrating the accuracy of the automated localization.

\begin{table}[ht]
  \caption{Performance comparison of different segmentation models for liver and tumor segmentation.}
  \centering
  \begin{tabular}{lc}
    \toprule
    \multicolumn{1}{c}{Model} & \multicolumn{1}{c}{Mean Dice} \\
    \midrule
    U-Net        & 0.9453 \\
    SegResNet    & 0.9416 \\
    DynUNet      & 0.9482 \\
    SwinUNETR    & \textbf{0.9516} \\
    \bottomrule
  \end{tabular}

  \label{tab:seg_performance}
\end{table}

\begin{table}[htbp]
  \caption{Reconstruction quality comparison between NeoGen VAE and QNeoGen VQVAE.}
  \centering
  \begin{tabular}{lccc}
    \toprule
    \multicolumn{1}{c}{Model}
      & \multicolumn{1}{c}{LPIPS $\downarrow$}
      & \multicolumn{1}{c}{SSIM $\uparrow$}
      & \multicolumn{1}{c}{PSNR $\uparrow$} \\
    \midrule
    QNeoGen VQVAE & 0.3003 & 0.4958 & 15.2210 \\
    NeoGen VAE & \textbf{0.2010} & \textbf{0.7408} & \textbf{22.7427} \\
    \bottomrule
  \end{tabular}

  \label{tab:vae_quality}
\end{table}

\begin{table}[htbp]
  \caption{Fréchet Inception Distance (FID) comparison across Axial (A), Sagittal (S), and Coronal (C) views.}
  \centering
  {\small
    \begin{tabular}{@{}lcccc@{}}
    \toprule
    \multicolumn{1}{c}{Method}
      & \multicolumn{1}{c}{FID $\downarrow$ (A)}
      & \multicolumn{1}{c}{FID $\downarrow$ (S)}
      & \multicolumn{1}{c}{FID $\downarrow$ (C)}
      & \multicolumn{1}{c}{FID $\downarrow$ (Avg.)} \\
    \midrule
    QNeoGen DM & 18.9751  & 22.1884 & 39.2260 & 26.7965 \\
    QNeoGen    & 14.5820  & 18.4123  & 28.6891 & 20.5611 \\
    NeoGen DM & 7.6284  & 10.8560 & 17.8180 & 12.1008 \\
    NeoGen     & \textbf{2.9440}  & \textbf{4.8279}  & \textbf{8.2223}  & \textbf{5.3314} \\
    \bottomrule
    \end{tabular}
  }

  \label{tab:fid_comparison}
\end{table}

\begin{figure}[ht]
\centering
\includegraphics[width=0.85\columnwidth]{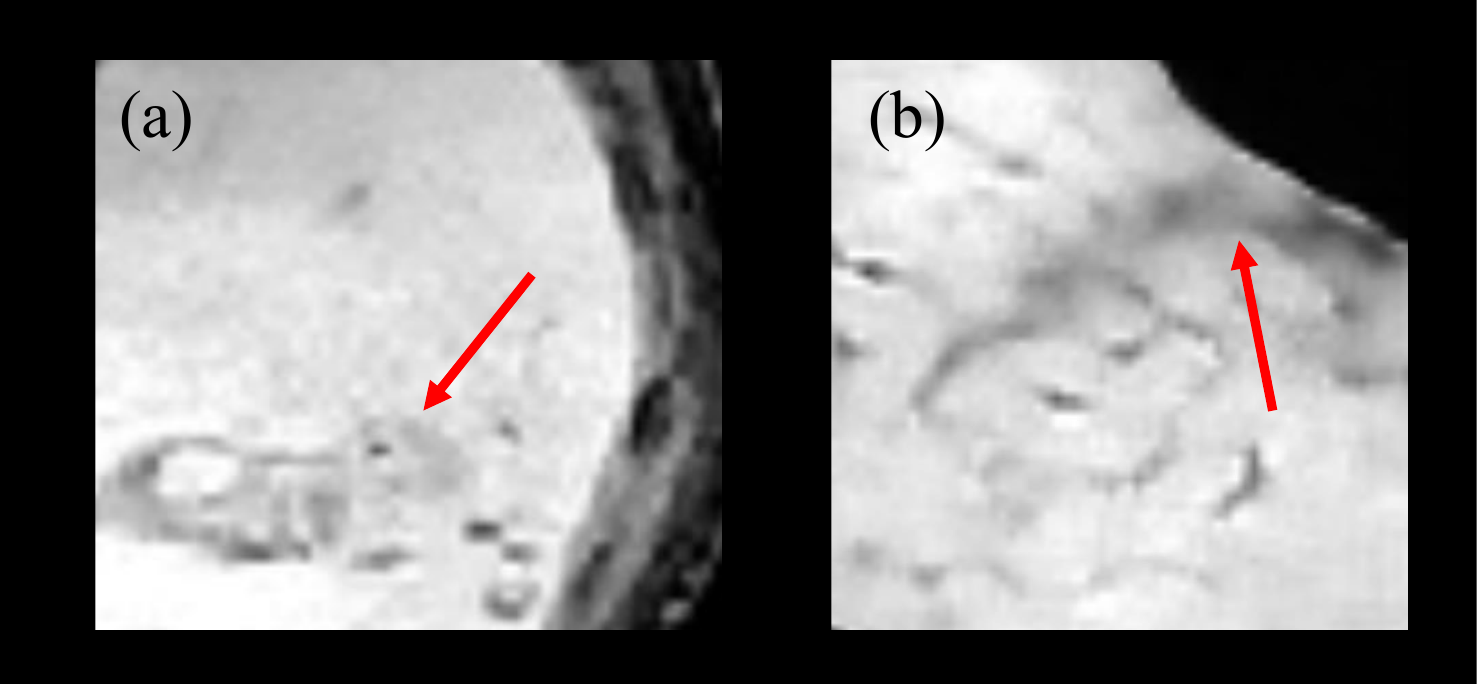}
\caption{Real Patch (a) and Synthetic Patch (b) generated by NeoGen using the corresponding anatomical mask. Morphological features are highlighted by red arrows.}
\label{fig6}
\end{figure}

\begin{table*}[ht]
\caption{AUC comparison of PNI classifiers trained on different R (Real) to S (Synthetic) ratios using 5-fold cross-validation. R+S columns represent progressive augmentation towards a fully balanced (1:1) dataset (R+S 100\% Balanced). Results are presented as Mean AUC. All baseline models are 3D implementations.}
\centering
\resizebox{\textwidth}{!}{%
\begin{tabular}{@{}lccccc@{}}
\toprule
\multicolumn{1}{c}{Model (3D)} & R (Imbalanced) & R + S (25\%) & R + S (50\%) & R + S (75\%) & R + S (100\% Balanced) \\ \midrule
ResNet-50 & 0.6938 & 0.7388 & 0.7599 & 0.7551 & 0.7618 \\
ResNet-101 & 0.6284 & 0.6989 & 0.7024 & 0.7117 & 0.7203 \\
ResNet-152 & \textbf{0.7078} & 0.7233 & 0.7421 & 0.7382 & 0.7526 \\
ResNet-200 & 0.6687 & 0.7250 & 0.7307 & 0.7397 & 0.7412 \\
DenseNet-121 & 0.6788 & 0.7443 & 0.7380 & 0.7490 & 0.7551 \\
DenseNet-169 & 0.6725 & 0.7021 & 0.7387 & 0.7434 & 0.7584 \\
DenseNet-201 & 0.6911 & 0.7133 & 0.7226 & 0.7278 & 0.7311 \\
DenseNet-264 & 0.6712 & 0.6899 & 0.7012 & 0.7037 & 0.7155 \\
EfficientNet-B0 & 0.6798 & 0.6897 & 0.6995 & 0.6910 & 0.6998 \\
EfficientNet-B1 & 0.6725 & 0.7127 & 0.7098 & 0.7101 & 0.7186 \\
EfficientNet-B2 & 0.6623 & 0.6899 & 0.7085 & 0.7024 & 0.7110 \\
EfficientNet-B3 & 0.6533 & 0.7215 & 0.7462 & 0.7671 & 0.7755 \\
SwinTransformer & 0.6829 & 0.7423 & 0.7454 & 0.7503 & 0.7522 \\
\textbf{PattenNet (Ours)} & 0.7001 & \textbf{0.7577} & \textbf{0.7670} & \textbf{0.7812} & \textbf{0.7903} \\ \midrule
Mean AUC & 0.6760 & 0.7178 & 0.7294 & 0.7336 & 0.7417 \\ \bottomrule
\end{tabular}%
}
\label{tab:auc_comparison}
\end{table*}

\subsection{Synthetic MRI Generation Quality}
We evaluated the quality of synthetic images generated by NeoGen. Figure~\ref{fig6} compares a real patch with a synthetic patch generated using the corresponding anatomical mask, demonstrating the fidelity of the generated morphology.

\subsubsection{Reconstruction Quality}
We compared the standard VAE used in NeoGen against a Vector Quantized VAE (VQVAE) variant (QNeoGen). As shown in Table~\ref{tab:vae_quality}, NeoGen VAE demonstrated superior performance across all metrics (LPIPS, SSIM, PSNR).

\subsubsection{Generation Fidelity}
Table~\ref{tab:fid_comparison} compares the FID scores. NeoGen (3D LDM + ControlNet with VAE) achieved the best average FID (5.3314), significantly outperforming the baseline 3D LDM (NeoGen DM, 12.1008) and the QNeoGen variants.

\subsection{PNI Prediction Performance}
Table~\ref{tab:auc_comparison} summarizes the PNI prediction performance of various 3D classification models. We evaluated performance using the original imbalanced real data (R) and augmented datasets where synthetic data (S) was progressively added to achieve a fully balanced (1:1) dataset.

\subsubsection{Impact of Synthetic Data Augmentation and Balancing}
The inclusion of synthetic data generated by NeoGen consistently improved performance across all models. When trained on the original imbalanced real data only (R), the mean AUC across all models was 0.6760. When the dataset was fully balanced using synthetic data (R+S 100\% Balanced), the mean AUC boosted to 0.7417.

\subsubsection{Model Comparison}
Our proposed PattenNet consistently outperformed all baseline 3D architectures, including established models such as ResNet~\cite{He2016}, DenseNet~\cite{Huang2017}, EfficientNet~\cite{Tan2019}, and SwinTransformer~\cite{Liu2021}. When trained on real data only, PattenNet achieved an AUC of 0.7001. With the fully balanced dataset, PattenNet achieved the highest AUC of 0.7903. This highlights the effectiveness of the specialized dual attention mechanism and the use of the frozen LDM encoder for PNI detection.

\begin{table}[ht]
    \caption{Ablation study evaluating the impact of attention components and the number of Dual Attention Blocks (DABs) in PattenNet.}
    \centering
    \begin{tabular}{lc}
    \toprule
    \multicolumn{1}{c}{Configuration} & \multicolumn{1}{c}{AUC (R+S)} \\
    \midrule
    \textbf{PattenNet (Full Model, 2 DABs)} & \textbf{0.7903} \\
    \midrule
    \textit{DAB Component Ablation (2 Blocks)} & \\
    w/ Channel Attention only & 0.7276 \\
    w/ Spatial Attention only & 0.7410 \\
    \midrule
    \textit{Number of DABs} & \\
    0 Blocks (Base LDM Encoder) & 0.7001 \\
    1 Block & 0.7589 \\
    3 Blocks & 0.7713 \\
    \bottomrule
    \end{tabular}
    \label{tab:ablation_study}
\end{table}

\section{Ablation Study}
We conducted ablation studies to evaluate the contribution of key components in the PattenNet architecture (Table~\ref{tab:ablation_study}). All ablation experiments were conducted using the R+S (100\% Balanced) data regime.

\subsubsection{Impact of Attention Mechanisms}
We first evaluated the importance of the specific components within the DABs in the proposed 2-block architecture. Removing both channel and spatial attention (equivalent to 0 DABs, AUC 0.7001) significantly degraded performance. Using only channel attention (AUC 0.7276) or only spatial attention (AUC 0.7410) provided substantially inferior performance compared to the full configuration (AUC 0.7903). This confirms that the synergistic integration of both channel and spatial attention mechanisms is crucial for capturing the complex features of PNI.

\subsubsection{Impact of the Number of DABs}
We investigated the optimal depth of the attention mechanism by varying the number of stacked DABs from 0 to 3. The configuration with 0 blocks (Base LDM Encoder) yielded the lowest AUC of 0.7001. Adding a single DAB significantly improved performance to 0.7589. The performance peaked with two DABs (AUC 0.7903). Increasing the depth further to three DABs resulted in a decrease in performance (AUC 0.7713), suggesting that two blocks provide the optimal balance of feature refinement for this task and dataset size.

\subsubsection{Limitations}
Despite the promising results, this study has limitations. The PNI prediction cohort is relatively small and sourced from a single institution, which may limit generalizability. External validation using multi-center datasets is required. Furthermore, the current framework relies on single-phase (hepatobiliary) MRI. Integrating multi-phase MRI data could provide complementary information.

\section{Conclusion}
We proposed NeoNet, an end-to-end 3D MRI-based deep learning framework for non-invasive prediction of perineural invasion (PNI). Our framework combines three modules: automated tumor localization (NeoSeg), conditioned data generation(NeoGen), and an optimized PNI prediction Network (PattenNet). This integration addresses the fundamental challenges of subtle PNI features and data scarcity inherent to PNI research.
 Our results demonstrate that anatomically constrained synthetic data augmentation enhances performance, and PattenNet achieves a superior AUC of 0.7903 for PNI prediction in cholangiocarcinoma. This work advances non-invasive diagnostic capabilities, enabling improved preoperative risk assessment and individualized treatment planning.

\section{Acknowledgement}
This work was supported by the Next Generation Semiconductor Convergence and Open Sharing System, by the Institute of Information \& Communications Technology Planning \& Evaluation (IITP) under the Artificial Intelligence Semiconductor Support Program to Nurture the Best Talents (IITP-2023-RS-2023-00256081), and by the High-Performance Computing Support Project, all funded by the Korea government (Ministry of Science and ICT).

\bibliographystyle{splncs04}
\bibliography{references}

@article{Liebig2009,
  title={Perineural invasion in cancer: a review of the literature},
  author={Liebig, Catherine and Ayala, Gustavo and Wilks, Jonathan A and Berger, David H and Albo, Daniel},
  journal={Cancer: Interdisciplinary International Journal of the American Cancer Society},
  volume={115},
  number={15},
  pages={3379--3391},
  year={2009},
  publisher={Wiley Online Library}
}

@article{Amit2016,
  title={Role of the nervous system in cancers: a review},
  author={Wang, Huan and Zheng, Qiming and Lu, Zeyi and Wang, Liya and Ding, Lifeng and Xia, Liqun and Zhang, Hao and Wang, Mingchao and Chen, Yicheng and Li, Gonghui},
  journal={Cell death discovery},
  volume={7},
  number={1},
  pages={76},
  year={2021},
  publisher={Nature Publishing Group UK London}
}

@book{Hruban2022,
  title={WHO classification of tumours of the digestive system.},
  author={Bosman, Fred T and Carneiro, Fatima and Hruban, Ralph H and Theise, Neil D},
  number={Ed. 4},
  year={2010}
}

@article{Qian2024,
  title={Neuroendocrine regulation of cholangiocarcinoma: A status quo review},
  author={Sha, Meng and Cao, Jie and Sun, Han-yong and Tong, Ying and Xia, Qiang},
  journal={Biochimica et Biophysica Acta (BBA)-Reviews on Cancer},
  volume={1872},
  number={1},
  pages={66--73},
  year={2019},
  publisher={Elsevier}
}

@article{Purohit2019,
  title={Imaging of perineural tumor spread in head and neck cancer},
  author={Ginsberg, Lawrence E},
  booktitle={Seminars in Ultrasound, CT and MRI},
  volume={20},
  number={3},
  pages={175--186},
  year={1999},
  organization={Elsevier}
}

@article{Gillies2016,
  title={Radiomics: images are more than pictures, they are data},
  author={Gillies, Robert J and Kinahan, Paul E and Hricak, Hedvig},
  journal={Radiology},
  volume={278},
  number={2},
  pages={563--577},
  year={2016},
  publisher={Radiological Society of North America}
}

@article{Lambin2017,
  title={Radiomics: the bridge between medical imaging and personalized medicine},
  author={Lambin, Philippe and Leijenaar, Ralph TH and Deist, Timo M and Peerlings, Jurgen and De Jong, Evelyn EC and Van Timmeren, Janita and Sanduleanu, Sebastian and Larue, Ruben THM and Even, Aniek JG and Jochems, Arthur and others},
  journal={Nature reviews Clinical oncology},
  volume={14},
  number={12},
  pages={749--762},
  year={2017},
  publisher={Nature Publishing Group UK London}
}

@article{Huang2021,
  title={Feasibility of magnetic resonance imaging-based radiomics features for preoperative prediction of extrahepatic cholangiocarcinoma stage},
  author={Huang, Xinqiao and Shu, Jian and Yan, Yulan and Chen, Xin and Yang, Chunmei and Zhou, Tiejun and Li, Man},
  journal={European Journal of Cancer},
  volume={155},
  pages={227--235},
  year={2021},
  publisher={Elsevier}
}

@article{Qi2025,
  title={An MRI-based fusion model for preoperative prediction of perineural invasion status in patients with intrahepatic cholangiocarcinoma},
  author={Qi, Zuochao and Yuan, Hao and Li, Qingshan and Chen, Pengyu and Li, Dongxiao and Chen, Kunlun and Meng, Bo and Ning, Peigang and Yu, Haibo and Li, Deyu},
  journal={World Journal of Surgical Oncology},
  volume={23},
  number={1},
  pages={164},
  year={2025},
  publisher={Springer}
}

@inproceedings{Kingma2013,
  title={Auto-encoding variational bayes},
  author={Kingma, Diederik P and Welling, Max},
  journal={arXiv preprint arXiv:1312.6114},
  year={2013}
}

@inproceedings{Goodfellow2014,
  title={Generative adversarial nets},
  author={Goodfellow, Ian and Pouget-Abadie, Jean and Mirza, Mehdi and Xu, Bing and Warde-Farley, David and Ozair, Sherjil and Courville, Aaron and Bengio, Yoshua},
  booktitle={Advances in Neural Information Processing Systems (NeurIPS)},
  volume={27},
  year={2014}
}

@article{Sohl-Dickstein2015,
  title={Deep unsupervised learning using nonequilibrium thermodynamics},
  author={Sohl-Dickstein, Jascha and Weiss, Eric and Maheswaranathan, Niru and Ganguli, Surya},
  journal={Proceedings of the International Conference on Machine Learning (ICML)},
  pages={2256--2265},
  year={2015}
}

@inproceedings{Ho2020,
  title={Denoising diffusion probabilistic models},
  author={Ho, Jonathan and Jain, Ajay and Abbeel, Pieter},
  booktitle={Advances in Neural Information Processing Systems (NeurIPS)},
  volume={33},
  pages={6840--6851},
  year={2020}
}

@inproceedings{Rombach2022,
  title={High-resolution image synthesis with latent diffusion models},
  author={Rombach, Robin and Blattmann, Andreas and Lorenz, Dominik and Esser, Patrick and Ommer, Bj{\"o}rn},
  booktitle={Proceedings of the IEEE/CVF conference on computer vision and pattern recognition (CVPR)},
  pages={10684--10695},
  year={2022}
}

@inproceedings{Zhang2023,
  title={Adding conditional control to text-to-image diffusion models},
  author={Zhang, Lvmin and Rao, Anyi and Agrawala, Maneesh},
  booktitle={Proceedings of the IEEE/CVF International Conference on Computer Vision (ICCV)},
  pages={3836--3847},
  year={2023}
}

@article{Yi2019,
  title={Generative adversarial network in medical imaging: A review},
  author={Yi, Xin and Walia, Ekta and Babyn, Paul},
  journal={Medical image analysis},
  volume={58},
  pages={101552},
  year={2019}
}

@article{Kazeminia2020,
  title={GANs for medical image analysis},
  author={Kazeminia, Somayeh and Baur, Christoph and Kuijper, Arjan and van Ginneken, Bram and Navab, Nassir and Albarqouni, Shadi and Mukhopadhyay, Anirban},
  journal={Artificial intelligence in medicine},
  volume={109},
  pages={101938},
  year={2020}
}

@inproceedings{Pinaya2022,
    title={Brain imaging generation with latent diffusion models},
    author={Pinaya, Walter H. L. and Tudosiu, Petru-Daniel and Gray, Robert and Rees, Geraint and Cardoso, M. Jorge and Ourselin, Sebastien and Barkhof, Frederik},
    booktitle={International Workshop on Simulation and Synthesis in Medical Imaging (SASHIMI)},
    pages={117--126},
    year={2022},
    organization={Springer}
}

@inproceedings{Ronneberger2015,
  title={U-net: Convolutional networks for biomedical image segmentation},
  author={Ronneberger, Olaf and Fischer, Philipp and Brox, Thomas},
  booktitle={Medical Image Computing and Computer-Assisted Intervention (MICCAI)},
  pages={234--241},
  year={2015},
  organization={Springer}
}

@article{Myronenko2018,
  title={3D MRI brain tumor segmentation using autoencoder regularization},
  author={Myronenko, Andriy},
  journal={Brainlesion: Glioma, Multiple Sclerosis, Stroke and Traumatic Brain Injuries},
  pages={311--320},
  year={2018},
  publisher={Springer}
}

@article{Isensee2021,
  title={nnU-Net: a self-configuring method for deep learning-based biomedical image segmentation},
  author={Isensee, Fabian and Jaeger, Paul F and Kohl, Simon AA and Petersen, Jens and Maier-Hein, Klaus H},
  journal={Nature Methods},
  volume={18},
  number={2},
  pages={203--211},
  year={2021}
}

@inproceedings{Tang2022,
  title={Self-supervised pre-training of swin transformers for 3d medical image analysis},
  author={Tang, Yucheng and Yang, Dong and Li, Wenqi and Roth, Holger R and Landman, Bennett and Xu, Daguang and Nath, Vishwesh and Hatamizadeh, Ali},
  booktitle={Proceedings of the IEEE/CVF conference on computer vision and pattern recognition},
  pages={20730--20740},
  year={2022}
}

@inproceedings{He2016,
  title={Deep residual learning for image recognition},
  author={He, Kaiming and Zhang, Xiangyu and Ren, Shaoqing and Sun, Jian},
  booktitle={Proceedings of the IEEE conference on computer vision and pattern recognition (CVPR)},
  pages={770--778},
  year={2016}
}

@inproceedings{Huang2017,
  title={Densely connected convolutional networks},
  author={Huang, Gao and Liu, Zhuang and Van Der Maaten, Laurens and Weinberger, Kilian Q},
  booktitle={Proceedings of the IEEE conference on computer vision and pattern recognition (CVPR)},
  pages={4700--4708},
  year={2017}
}

@inproceedings{Tan2019,
  title={Efficientnet: Rethinking model scaling for convolutional neural networks},
  author={Tan, Mingxing and Le, Quoc V},
  booktitle={International conference on machine learning (ICML)},
  pages={6105--6114},
  year={2019},
  organization={PMLR}
}

@inproceedings{Vaswani2017,
  title={Attention is all you need},
  author={Vaswani, Ashish and Shazeer, Noam and Parmar, Niki and Uszkoreit, Jakob and Jones, Llion and Gomez, Aidan N and Kaiser, {\L}ukasz and Polosukhin, Illia},
  booktitle={Advances in neural information processing systems (NeurIPS)},
  volume={30},
  year={2017}
}

@inproceedings{Dosovitskiy2020,
  title={An image is worth 16x16 words: Transformers for image recognition at scale},
  author={Dosovitskiy, Alexey and Beyer, Lucas and Kolesnikov, Alexander and Weissenborn, Dirk and Zhai, Xiaohua and Unterthiner, Thomas and Dehghani, Mostafa and Minderer, Matthias and Heigold, Georg and Gelly, Sylvain and others},
  booktitle={International Conference on Learning Representations (ICLR)},
  year={2020}
}

@inproceedings{Liu2021,
  title={Swin transformer: Hierarchical vision transformer using shifted windows},
  author={Liu, Ze and Lin, Yutong and Cao, Yue and Hu, Han and Wei, Yixuan and Zhang, Zheng and Lin, Stephen and Guo, Baining},
  booktitle={Proceedings of the IEEE/CVF international conference on computer vision (ICCV)},
  pages={10012--10022},
  year={2021}
}

@inproceedings{Woo2018,
  title={Cbam: Convolutional block attention module},
  author={Woo, Sanghyun and Park, Jongchan and Lee, Joon-Young and Kweon, In So},
  booktitle={Proceedings of the European conference on computer vision (ECCV)},
  pages={3--19},
  year={2018}
}

@inproceedings{Heusel2017,
  title={Gans trained by a two time-scale update rule converge to a local nash equilibrium},
  author={Heusel, Martin and Ramsauer, Hubert and Unterthiner, Thomas and Nessler, Bernhard and Hochreiter, Sepp},
  booktitle={Advances in Neural Information Processing Systems (NeurIPS)},
  volume={30},
  year={2017}
}

@inproceedings{Loshchilov2017,
  title={Decoupled weight decay regularization},
  author={Loshchilov, Ilya and Hutter, Frank},
  booktitle={International Conference on Learning Representations (ICLR)},
  year={2017}
}

@inproceedings{Zhang2018,
  title={The unreasonable effectiveness of deep features as a perceptual metric},
  author={Zhang, Richard and Isola, Phillip and Efros, Alexei A and Shechtman, Eli and Wang, Oliver},
  booktitle={Proceedings of the IEEE conference on computer vision and pattern recognition (CVPR)},
  pages={586--595},
  year={2018}
}

\end{document}